\documentclass[11pt]{article}

\usepackage[preprint]{acl}

\usepackage{booktabs}
\usepackage{tabularx}
\usepackage{footnote}
\usepackage{siunitx} 
\usepackage{times}
\usepackage{latexsym}
\usepackage{graphicx}
\usepackage{booktabs}
\usepackage{tabularx}
\usepackage{amsmath, bm}
\usepackage{amsfonts}
\usepackage{multirow, multicol}
\usepackage{tabularx}
\usepackage{makecell}
\usepackage{array} 
\usepackage[most]{tcolorbox}
\usepackage{listings}
\usepackage{fancyvrb}

\usepackage{colortbl}
\usepackage[table]{xcolor} 
\definecolor{mygray}{gray}{.9}

\usepackage[T1]{fontenc}

\usepackage[utf8]{inputenc}

\usepackage{microtype}

\usepackage{inconsolata}

\usepackage{graphicx}

\title{MMViR: A Multi-Modal and Multi-Granularity Representation for Long-range Video Understanding}

\author{
  {\bf Zizhong Li}$^\ast$, {\bf Haopeng Zhang}$^\dagger$, {\bf Jiawei Zhang}$^\ast$ \\
  $^\ast$IFM Lab, University of California, Davis \\
  $^\dagger$ALOHA Lab, University of Hawaii at Mānoa \\
  \texttt{zzoli@ucdavis.edu, haopengz@hawaii.edu, jiawei@ifmlab.org}
}

\begin{document}
\maketitle
\begin{abstract}
Long videos, ranging from minutes to hours, present significant challenges for current Multi-modal Large Language Models (MLLMs) due to their complex events, diverse scenes, and long-range dependencies. 
Direct encoding of such videos is computationally too expensive, while simple video-to-text conversion often results in redundant or fragmented content.
To address these limitations, we introduce \textbf{MMViR}, a novel multi-modal, multi-grained structured representation for long video understanding. MMViR identifies key turning points to segment the video and constructs a three-level description that couples global narratives with fine-grained visual details. This design supports efficient query-based retrieval and generalizes well across various scenarios. 
Extensive evaluations across three tasks, including QA, summarization, and retrieval, show that MMViR outperforms the prior strongest method, achieving a 19.67\% improvement in hour-long video understanding while reducing processing latency to 45.4\% of the original.
\end{abstract}

\section{Introduction}
\label{sec1}
    Over the past few years, Large Language Models (LLMs) have shown remarkable progress, achieving state-of-the-art performance on various NLP tasks \cite{zhang2023extractive, zhang2023summit, yao2024survey, li2024survey, wu2025survey}.
    Building on this success, recent research has extended LLMs to the multi-modal domain, giving rise to Multi-modal Large Language Models (MLLMs), which broaden language understanding to visual reasoning and advance the capabilities of vision-language systems \cite{yin2024survey, zhang2024mm, zhang2025bridging}.
    
    Despite their strong performance on image and short-video tasks, MLLMs remain fundamentally limited when processing long-range video content spanning tens of minutes to over an hour \cite{liang2024survey}.
    Most current MLLMs \cite{zhang2023videollamainstructiontunedaudiovisuallanguage, qwen2025qwen25technicalreport} are restricted to process a limited number of frames due to constraints in computational complexity and architecture scalability. 
    \begin{figure}
        \centering
        \includegraphics[width=1.0\linewidth]{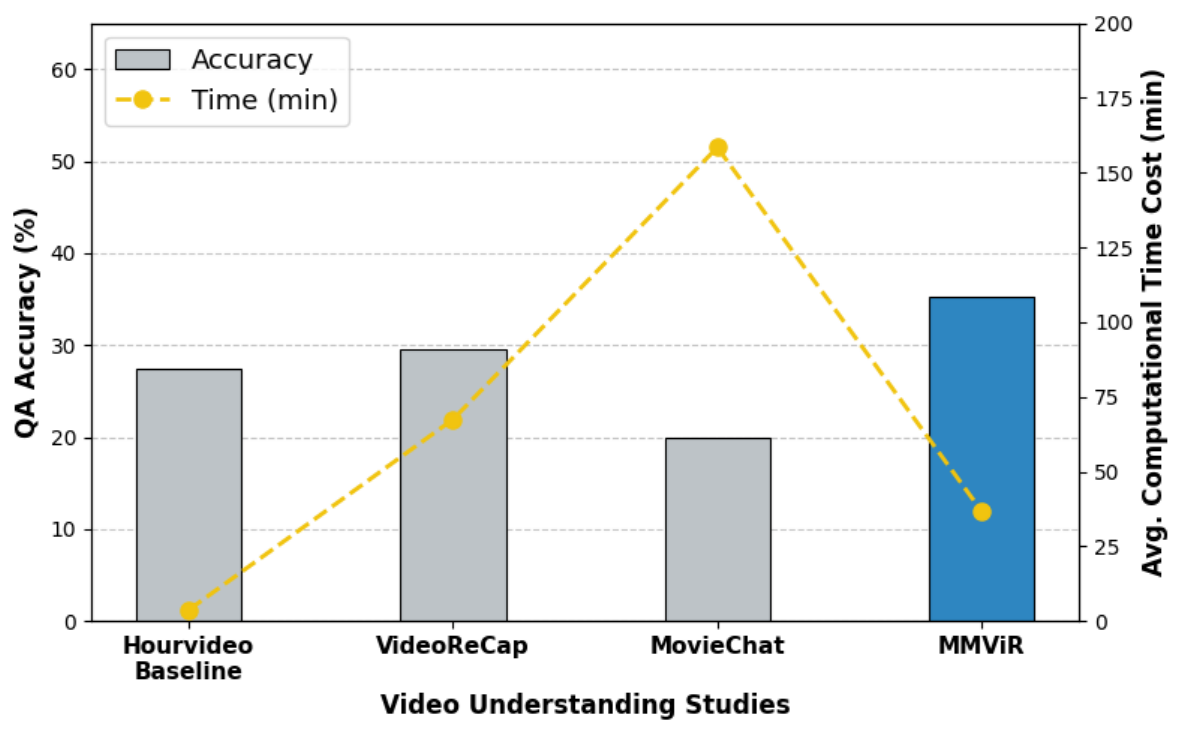}
        \caption{Performance comparison on Hourvideo QA benchmark. MMViR consistently outperforms prior methods by simultaneously optimizing for reasoning precision and inference efficiency.}
        \label{fig:01}
        \vspace{-4mm}
    \end{figure}
    To mitigate these issues, prior research has explored two main directions. One is to enhance MLLM architectures with specialized modules, such as long/short-term memory \cite{song2024moviechat+, he2024malmm} or redundancy-reduction mechanisms \cite{li2025videochatflashhierarchicalcompressionlongcontext, wang2025adaretake}.
    However, this approach still struggles to scale to extremely long videos, where the computational cost of dense frame processing becomes prohibitive.
    An alternative strategy is to transcribe long-range video content into textual representation, either by aggregating clip textual descriptions \cite{chandrasegaran2024hourvideo, pang2025mr} or multi-level reasoning chains \cite{sanders2024tv}.
    Nevertheless, such representations often suffer from content fragmentation that undermines global narrative coherence and overlooks critical details.
    
    This persistent trade-off leads us to rethink the fundamental design of video representations, specifically by focusing on two key dimensions: 1) \textbf{Structure: } identifying the optimal way to transform long video content without sacrificing key information; and 2) \textbf{Modality: }exploring whether incorporating additional modalities beyond textual descriptions can represent video information more effectively.
    Using VideoQA as a benchmark, we evaluate various representation strategies by analyzing the trade-off between accuracy and token consumption.
    This analysis leads to two key findings. 
    First, \textbf{multi-grained structures}, spanning from high-level summaries to fine-grained details, provide a more coherent way to present long videos. Second, \textbf{multi-modal representations} that fuse text with visual representation offer better compression efficiency than unimodal approaches.

    Based on these findings, we introduce MMViR, a Multi-modal and Multi-grained Video Representation designed for long-video understanding.
    MMViR organizes long videos into a structured, searchable multi-grained format with three distinct levels:
    1) a global timeline that captures the overall narrative flow; 2) clip-level coarse-grained textual descriptions from multiple perspectives to refine the timeline; 3) fine-grained visual representations paired with concise text for precise grounding.
    This structure ensures comprehensive coverage while enabling efficient retrieval of task-relevant information.
    As a result, the downstream tasks can access only the necessary content via MMViR, thereby significantly reducing computational overhead.
    
    We evaluate MMViR on three core long-video tasks (i.e., VideoQA, summarization, and retrieval), using datasets of varying temporal scales. Our experiments compare MMViR against several prior methods, and the results show that it consistently achieves better accuracy while maintaining higher computational efficiency.
    These findings, illustrated in Figure \ref{fig:01}, confirm the advantages of our approach for long-range video reasoning. In summary, our main contributions are:
    \begin{itemize}
        \item We conduct an in-depth analysis of the trade-offs between efficiency and accuracy in long-video understanding, identify key factors for effective representation, and demonstrate how to implement these factors in practice.
        \item We introduce MMViR, a multi-modal and multi-grained framework that bridges global coherence with fine-grained detail, enabling efficient query-based retrieval across diverse long-video tasks.
        \item Extensive experiments on various datasets and tasks show that MMViR outperforms previous methods in both efficiency and accuracy, particularly on hour-long video benchmarks.
    \end{itemize}

\section{Related Work}
\begin{figure}
    \centering
    \includegraphics[width=1.0\linewidth]{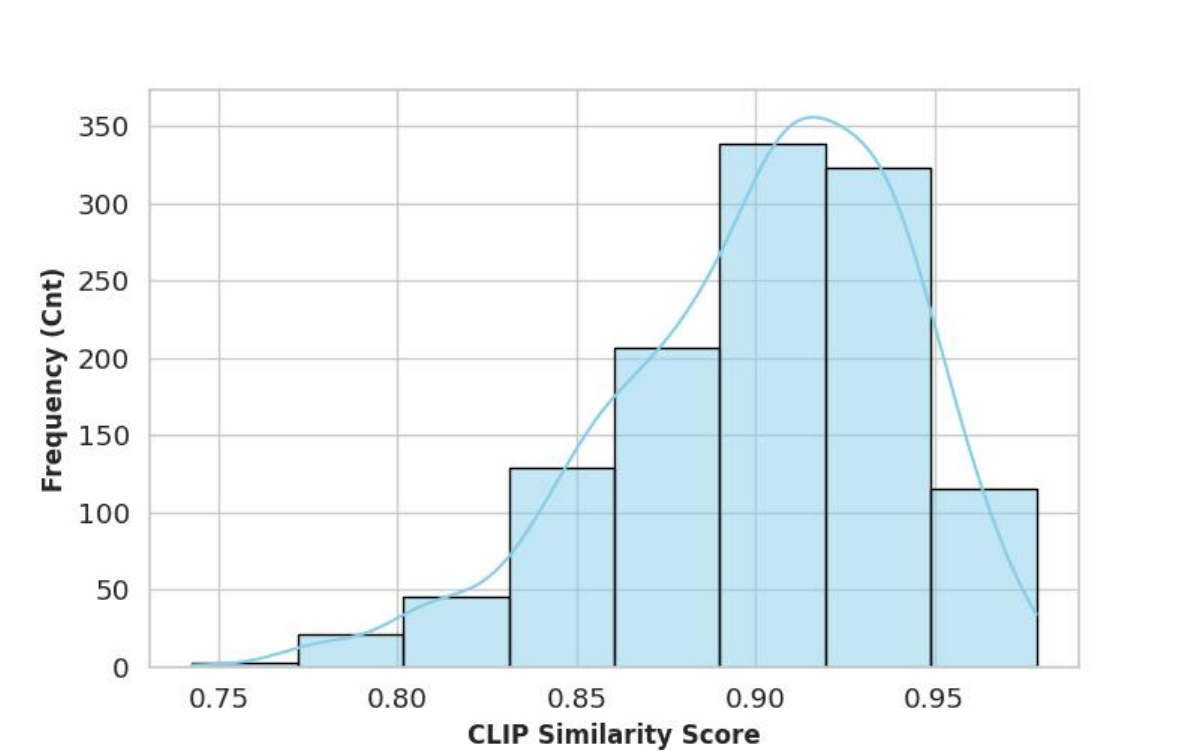}
    \caption{Overall distribution of frame-to-frame CLIP Similarity Scores of a sampled hour-long video.}
    \label{fig:clip-2}
\end{figure}
\subsection{Long Video Understanding}
Video understanding has long been a core challenge in the vision domain.
Early research focused on short-term action recognition and classification \cite{soomro2012ucf101, reddy2013recognizing, kay2017kinetics, carreira2017quo} using 2D CNNs with RNNs (e.g., LSTMs) \cite{donahue2015long}, or 3D CNNs \cite{taylor2010convolutional, ji20123d, tran2015learning} to capture temporal information. While effective for short clips, these models could not achieve a full understanding of video content.
Later, the introduction of longer video datasets \cite{rohrbach2015dataset, xu2016msr, zhou2018towards} shifted the focus toward text–video alignment, leading to tasks like video description \cite{yao2015describing, venugopalan2015sequence, pan2016hierarchical} and summarization \cite{zhao2017hierarchical, rochan2018video, zhang2018retrospective}.
However, these approaches often produced template-like outputs and missed fine-grained details \cite{li2019visual, apostolidis2021video}.

The advent of cross-modal pretraining, led by CLIP \cite{radford2021learning}, further advanced video understanding through models such as VideoCLIP \cite{xu2021videoclip}.
More recently, MLLMs \cite{zhang2023videollamainstructiontunedaudiovisuallanguage, qwen2025qwen25technicalreport} have integrated visual modalities into LLMs, yet they often struggle with hour-long videos due to restricted context windows.
Current efforts to address this, such as memory modules \cite{song2024moviechat+}, redundancy reduction \cite{li2025videochatflashhierarchicalcompressionlongcontext}, and text-based video conversion \cite{chandrasegaran2024hourvideo, pang2025mr}, only partially mitigate the problem and often incur high computational cost or produce incoherent information. 
These persistent gaps highlight the need for a structured, semantically faithful representation for long-video understanding.

\subsection{Video Representation}
Using language models to encode videos into structured representations has become an effective strategy for video understanding.
Earlier methods \cite{zhang2018cross, li2020hero, xiao2022hierarchical, ashutosh2023hiervl, cheng2024hico} typically trained language models on video–text pairs to build multi-level representations.
Recently, advances in MLLMs have enabled the generation of structured representations through multi-level captioning, in which models produce descriptions that serve as organized video summaries \cite{zala2023hierarchical, islam2024video, sanders2024tv, wu2025longvitu}.
However, most current methods of this type are query-specific, meaning they build representations tailored to a single question. This focus often limits global coherence and makes it difficult to generalize representation to other tasks.

In contrast, MMViR generates multi-grained, multi-modal representations in a query-agnostic manner, capturing both fine-grained details and the overall narrative. This design results in a coherent, retrieval-friendly structure that can be efficiently reused across diverse downstream tasks without being reconstructed for every new query.

\begin{figure}
    \centering
    \includegraphics[width=1.0\linewidth]{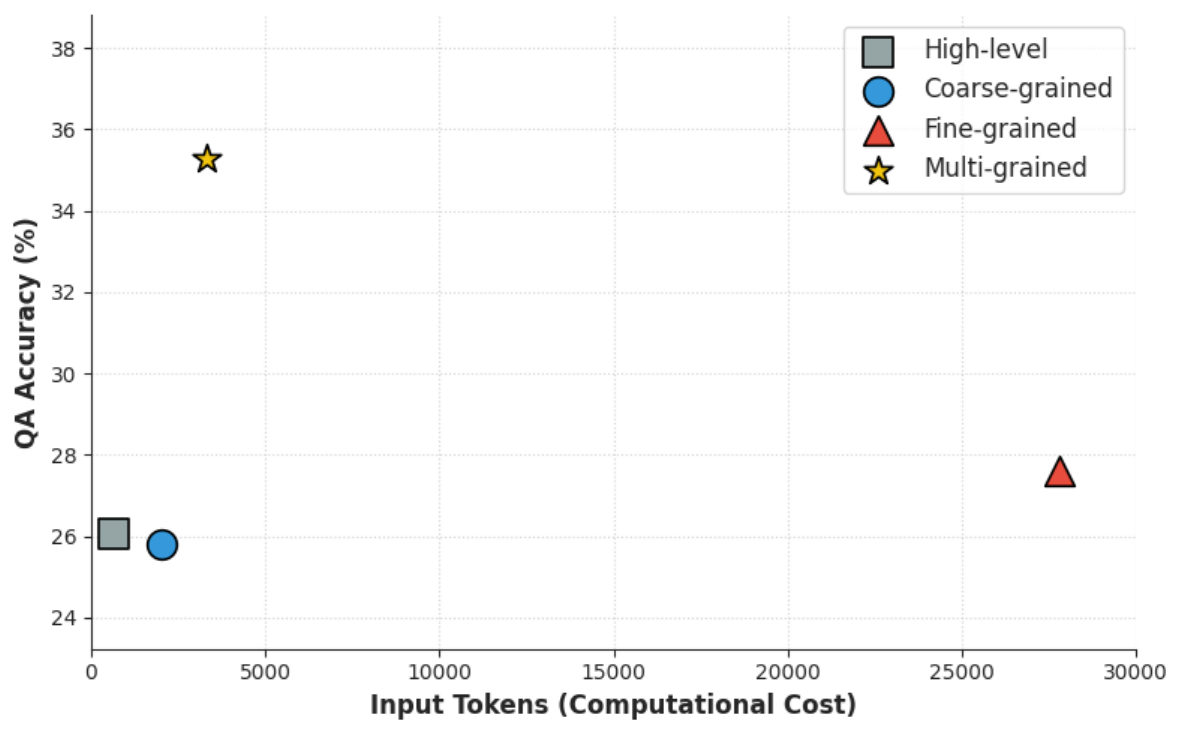}
    \caption{Token cost vs. QA accuracy for different granularities of video representations.}
    \label{fig:03}
    \vspace{-3mm}
\end{figure}

\begin{figure}
    \centering
    \includegraphics[width=1.0\linewidth]{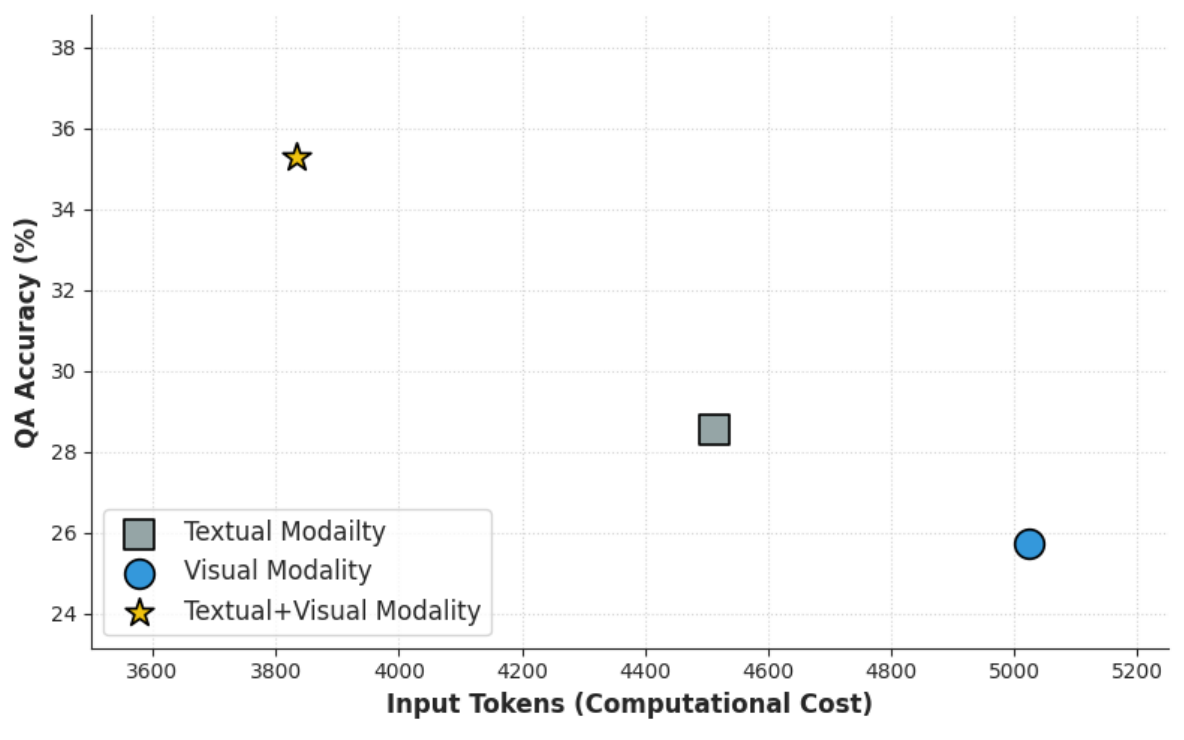}
    \caption{Token cost vs. QA accuracy for different modalities of video representations.}
    \label{fig:04}
    \vspace{-3mm}
\end{figure}
\begin{figure*}
    \centering
    \includegraphics[width=1.0\linewidth]{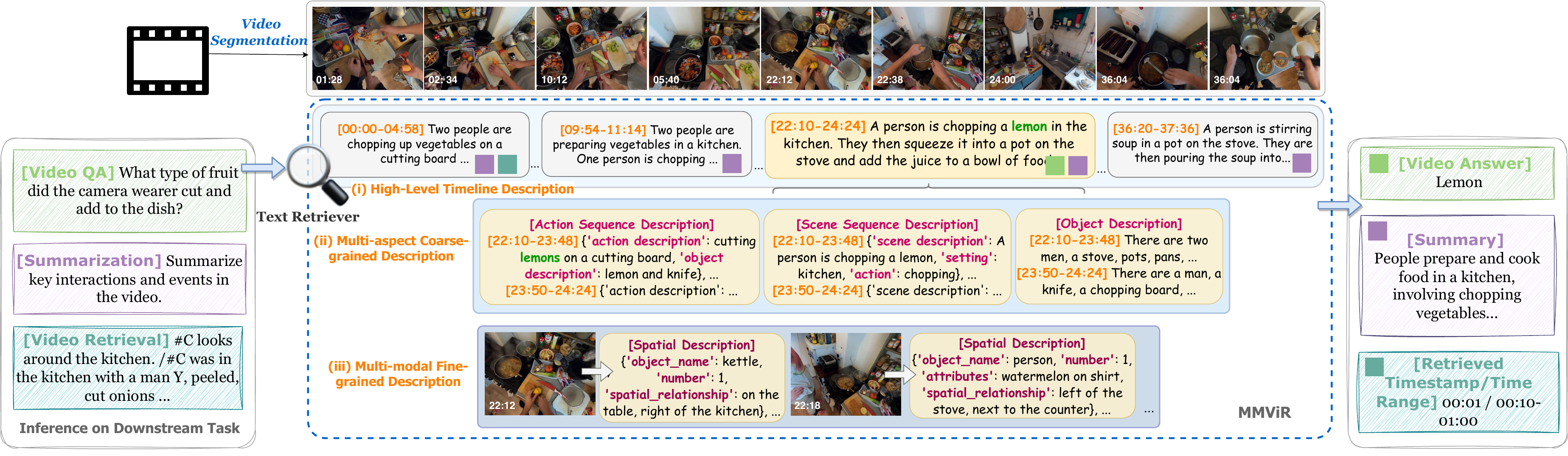}
    \caption{Overview of the MMViR and the downstream inference. Given a long video, MMViR leverages an MLLM to construct a multi-modal and multi-grained representation for it. The high-level timeline descriptions serve as a global semantic index, enabling efficient query-aware relevant clip localization. This design ensures a synergistic balance between global narrative coherence and fine-grained evidence, effectively addressing the computational challenges of long-form content.}
    \label{fig:main}
    \vspace{-5mm}
\end{figure*}
\section{Preliminary Investigation}
\label{sec:3}
In this section, we conduct a preliminary study to understand the intrinsic features of long video content. Instead of treating video as a flat sequence, we investigate its structure through three aspects: temporal coherence, information structure, and modal synergy.
These observations provide the foundation and design principles for our proposed video representation.

\subsection{Temporal Coherence of Long Video}
Long videos are inherently non-uniform, consisting of many events with varying durations.
While segmenting videos into fixed intervals is a common baseline to manage computational costs, this approach often disrupts the natural flow of information.
To investigate the intrinsic structure of long-form content, we analyzed frame-to-frame visual similarity in the \textit{Hourvideo} dataset \cite{chandrasegaran2024hourvideo}.
Specifically, we analyze 50 videos from it and sample at 0.5 fps, computing frame-to-frame similarity using CLIP \cite{radford2021learning}.
As Figure \ref{fig:clip-2} shows, frame similarities follow a "long-tailed" trend, indicating that \textbf{frame continuity is likely to be punctuated by these sparse similarity drops}, which serve as natural indicators of scene transitions (more details in Appendix \ref{sec:appendixA}). 
This semantic-aware segmentation, such as using "turning points" determined by CLIP or Kernel-based Temporal Segmentation (KTS) \cite{potapov2014category}, aligns more closely with the video's natural than fixed-interval splitting. 

\subsection{Impact of the Data Structure}

After determining \textit{where} to segment, we investigate what information should be preserved within each segment.
Drawing on insights from prior work \cite{zala2023hierarchical, sanders2024tv}, long-form reasoning often requires a way to \textit{zoom in and out}, locating an event across a duration and then analyzing its specific details.
We simulate this by generating textual descriptions at three level granularities via MLLM \cite{zhang2023videollamainstructiontunedaudiovisuallanguage}, including global summaries (i.e., high-level), action details (i.e., coarse-grained), and frame-specific descriptions (i.e., fine-grained).
By getting the response to 1,162 queries from \textit{Hourvideo} generated from GPT-4o \cite{openai2023gpt4}. For multi-grained inference, we first retrieve the top-10 most relevant clips using the global summaries, then provide the action details and frame description as context. 
We observe that relying on single granularity cannot achieve better performance than using a combination, as shown in Figure \ref{fig:03}.
\begin{figure}[htbp]
    \centering
    \includegraphics[width=0.98\linewidth]{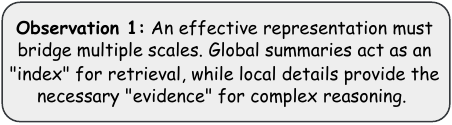}
    \vspace{-5mm}
\end{figure}

\subsection{Impact of Data Modality}
Finally, we examine whether textual descriptions alone are sufficient to represent these segments. While text is efficient for indexing, it often loses subtle visual information (e.g., specific spatial relationships).
We compare three configurations: text-only (i.e., multi-grained text description), vision-only (i.e., initial sampled frames), and a hybrid multi-modal approach.
As shown in Figure \ref{fig:04}, the combination representation substantially outperforms unimodal baselines, while also incurring lower computational overhead. 
\begin{figure}[htbp]
    \centering
    \includegraphics[width=0.98\linewidth]{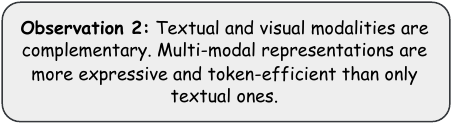}
    \vspace{-5mm}
\end{figure}

\section{Method}

\subsection{Overview of MMViR} 
Guided by the insights in Section \ref{sec:3}, we propose MMViR.
Unlike flat representations, MMViR explicitly models the multi-modal nature of long-form videos by organizing information into three levels of granularity:
1) \textit{High Level Global Timeline}: concise summaries of the dominant activity in each segmented clip; 2) \textit{Clip-level Coarse-grained Description}: multi-perspective details regarding actions, objects, and scenes content; 3) \textit{Fine-grained Description}: localized visual-textual pairs extracted from sampled frames for grounding.
As shown in Figure \ref{fig:main}, MMViR maintains a comprehensive record of video content while providing a retrieval-friendly format that generalizes across diverse downstream tasks without requiring query or task-specific reconstruction.

\subsection{Construction Pipeline of MMViR} 
\textbf{Event-based Video Segmentation.} 
Given a long video $\mathbb{V}$ spanning $[0, T]$, we first sample frames at a low fps (e.g., 0.5) and compute the CLIP \cite{radford2021learning} similarity between consecutive frames.
Following our observations of temporal non-uniformity in Section \ref{sec:3}, we detect the turning points $0=t_0<t_1<\dots<t_N=T$ within the similarity distribution (or via the KTS method \cite{potapov2014category}).
These points serve as indicators of potential scene changes to segment the video into $N$ coherent clips, denoted as $C_i=\mathbb{V}[t_{i-1},t_i]$ for $i=1,\dots,N$.\\
\textbf{Hierarchical Representation Generation.} 
For each segmented clip $C_i$, we utilize an MLLM \cite{zhang2023videollamainstructiontunedaudiovisuallanguage} to generate a multi-modal, multi-grained representation tuple $R(C_i)=(R^{timeline}(C_i), R^{coarse}(C_i), R^{fine}(C_i))$.
That is, the MLLM first generates a concise, one-sentence summary $s_{C_i}$ that captures the dominant activity and key events of the entire clip. The clip's overall activity and key events. This serves as the timeline-level representation:
\begin{equation}
    R^{timeline}(C_i)=\{s_{C_i}\},
\end{equation}

To achieve higher semantic precision at the coarse-grained level, we divide each clip $C_i$ into sub-segments of a fixed maximum duration.
This further segmentation prevents the MLLM from overlooking critical actions in longer clips.
For each sub-segment, the MLLM produces detailed descriptions $d(k)$ of the actions $A_k$, scenes $S_k$, and objects $O_k$:
\begin{equation}
    R^{coarse}(C_i) = \{d(k)=(A_k, S_k, O_k)\},
\end{equation}
where $A_k$, $S_k$, and $O_k$ denote the actions, scenes, and objects in the $k$-th sub-segment.

Finally, we sample frames within each sub-segment at a lower density. The MLLM pairs these with scene descriptions to form text-frame pairs $f_{k, m}$:
\begin{equation}
    R^{fine}(C_i)=\{f_{k, m}\mid m=1, \dots, M_k\}.
\end{equation}
where $M_k$ is the number of sampled frames in the $k$-th sub-segment.

\subsection{Query-based Retrieval and Reasoning}
MMViR's structured design enables an \textit{adaptive retrieval mechanism} tailored to specific task requirements.
For global tasks like summarization, we leverage the full $R^{timeline}$ to capture the overall narrative.
For localized reasoning tasks (e.g., VideoQA), we employ a two-stage process. In the first stage, a retrieval model (e.g., Contriever \cite{izacard2021contriever}) is used to retrieve the top-\textit{k} relevant clips using $R^{timeline}$ as the search index. 
Then, only the corresponding $R^{coarse}$ and $R^{fine}$ of the retrieved clips are expanded as content for the MLLM.
This mechanism allows the downstream inference to "zoom in" on localized evidence while retaining global context, maintaining high accuracy while significantly reducing the computational cost of long-video reasoning.

\section{Experiment}
\begin{table}[t] 
\centering
\footnotesize 
\setlength{\tabcolsep}{3pt} 
\begin{tabularx}{\columnwidth}{@{}l c c l X@{}}
\toprule
\textbf{Dataset} & \textbf{\#Video} & \textbf{\#Query} & \textbf{Duration} & \textbf{Tasks} \\ \midrule
HourVideo & 50 & 1,162 & 20-120m & VideoQA, Sum. \\
EgoSchema & 500 & 500 & $\sim$3m & VideoQA \\
Video-MME & 60 & 64 & $\sim$30m & VideoQA \\
MovieChat & 448 & 448
& 7-8m & Sum. \\
Ego4D & 50 & 36k/516
& 20-120m & Video Retrieval\\ 
\bottomrule
\end{tabularx}
\caption{Dataset statistics and evaluation tasks (Ego4D's query consists of 36k narrations and 516 summaries). }
\label{tab:dataset}
\vspace{-5mm}
\end{table}
We evaluate MMViR across five diverse long-form video datasets, covering three core tasks: VideoQA, summarization, and video retrieval. The results consistently validate the robustness and generality of our representation, highlighting MMViR's unique ability in balancing semantic completeness with computational efficiency across varying temporal scales.
\subsection{Experiment Setup}
\newcolumntype{Y}{>{\centering\arraybackslash}X} 
\newcolumntype{M}[1]{>{\centering\arraybackslash}m{#1}} 

\begin{table*}[ht]
\centering
\renewcommand{\arraystretch}{1.2}
\setlength{\tabcolsep}{1.1pt}       
\small
\resizebox{\textwidth}{!}{        
\begin{tabularx}{\textwidth}{c c c  Y Y Y Y}
\toprule
\multirow{2}*{\textbf{Method}} & \multirow{2}*{\textbf{Modality}} & \multirow{2}*{\textbf{Structure}} & 
\multirow{2}*{\textbf{Avg. Time (min) \boldsymbol{$\downarrow$}}} & 
\textbf{Acc} \boldsymbol{$\uparrow$} \textbf{(Hourvideo)} & \textbf{Acc \boldsymbol{$\uparrow$} (Egoschema) } & \textbf{Acc \boldsymbol{$\uparrow$} (VideoMME)} \\
\midrule
\rowcolor{gray!9} 
\multicolumn{7}{c}
{\textbf{\textsl{Directly Sample Frames}}}\\
\midrule
Uniform  & \multirow{1}*{Image}      & \multirow{1}*{Global}   & \multirow{1}*{--}  & \multirow{1}*{26.4}  & 60.4 & \textbf{54.7}  \\
Random  & \multirow{1}*{Image}       &  \multirow{1}*{Global}  & \multirow{1}*{--} & 26.6  & 61.8 & 48.4\\
\midrule
\rowcolor{gray!9} 
\multicolumn{7}{c}
{\textbf{\textsl{Comparable Baseline}}}\\
\midrule
Hourvideo (2024) & Text        & Fine-grained  & \underline{3.8} & 27.5 & 34.2 & 11.7 \\
VideoRecap (2024) & Text        & Multi-grained & 67.5 &  29.5 &  34.4 & 33.3 \\ 
MovieChat (2024) & Image  & Global    & 158.4 & 19.9 & 53.5 & 38.2 \\
\midrule
\rowcolor{gray!9} 
\multicolumn{7}{c}
{\textbf{\textsl{Ours (Kernel Temporal Segmentation)}}}\\
\midrule
MMViR (w/o img) & Text & Multi-grained & 33.1 &  25.2 & - & 42.2 \\
MMViR (w/o text)  & Image & Fine-grained & \textbf{0.8} & 28.2 & - & 48.4 \\
MMViR  & \multirow{1}*{Text+Image}  & \multirow{1}*{Multi-grained} & \multirow{1}*{33.9} & \multirow{1}*{30.5}  &  -  &  \multirow{1}*{46.9} \\
\midrule
\rowcolor{gray!9} 
\multicolumn{7}{c}
{\textbf{\textsl{Ours (Clip-based Segmentation)}}}\\
\midrule
MMViR (w/o img) & Text & Multi-grained & 36.0  & \underline{33.6} & 60.2 & 42.2 \\
MMViR (w/o text)  & Image & Fine-grained & \textbf{0.8} & 28.7 & \textbf{65.4} & \textbf{54.7} \\
MMViR  & \multirow{1}*{Text+Image}  & \multirow{1}*{Multi-grained} & \multirow{1}*{36.8} & \multirow{1}*{\textbf{35.3}}  &  \multirow{1}*{\underline{61.8}}  &  \multirow{1}*{\underline{51.7}} \\
\bottomrule
\end{tabularx}
}
\caption{Performance comparison on VideoQA tasks. The bold in the table indicates the best performance, while the underlined indicates the second best.}
\vspace{-2mm}
\label{tab:qa_comparison}
\end{table*}

\textbf{Dataset.} We evaluate MMViR on five benchmarks, with detailed statistics provided in Table \ref{tab:dataset}. \\
\textbf{1) Hourvideo} \cite{chandrasegaran2024hourvideo}: Our primary evaluation benchmark for extreme long-range video reasoning, providing hour-long videos paired with five-way multiple-choice QA pairs and summarization ground-truth.
\textbf{2) Egoschema} \cite{mangalam2023egoschema}: Used to evaluate the understanding of egocentric activities, where each question requires global information of the associated video content.
\textbf{3) Video-MME} \cite{fu2025video}: A benchmark provides a diverse set of six visual domains, which are used to evaluate the robustness across various content types of our representation in VideoQA.
\textbf{4) MovieChat-1K} \cite{song2024moviechat+}: Comprising diverse movies and TV series with manual annotations, specifically used to evaluate narrative summarization capabilities.
\textbf{5) Ego4D} \cite{grauman2022ego4d}: The benchmark offers high-density temporal narrations, which are sampled and are used to measure the efficiency and precision of video retrieval in long-form sequences.\\
\textbf{Evaluation Metrics. }To assess the effectiveness and efficiency of MMViR, we categorize our evaluation into performance- and cost-based metrics:\\
\textbf{1) VideoQA}. \textit{Accuracy}: The percentage of correctly answered questions. \textit{Time Consumption}: The average time (in minutes) to process the hour-long video (i.e., Hourvideo dataset) before QA. 
\textbf{2) Summarization. }The standard evaluation metrics are adopted, including ROUGE-2, ROUGE-L \cite{ng2015better}, and METEOR \cite{banerjee2005meteor}, to assess the quality of generated summaries.
\textbf{3) Video Retrieval. }The localization precision is assessed at two temporal scales: \textit{Frame-level Precision\@K: }Measuring the accuracy of pinpointing specific ground-truth frames. \textit{Range-level Overlap@K: }Evaluating the temporal intersection between retrieved segments and ground-truth intervals to measure temporal alignment.\\
\textbf{Baselines.} To evaluate MMViR against current state-of-the-art approaches, we select five representative baselines categorized by:\\
\textbf{1) Caption-based Representations:} \textit{Hourvideo Baseline} \cite{chandrasegaran2024hourvideo}, generates multi-aspect textual descriptions (actions, objects, and scenes) for fixed one-minute video segments. 
\textit{VideoRecap} \cite{islam2024video}, a recursive captioning framework that produces hierarchical descriptions at clip, segment, and video levels, with a focus on atomic human actions.
\textbf{2) Memory-based Architectures: }
\textit{MovieChat} \cite{song2024moviechat+}, a model that utilizes a long/short-term memory management mechanism. We evaluate it in its \textit{global mode} to process the entire video sequence end-to-end.
\textbf{3) End-to-End Multi-modal Retrievers: }
\textit{CLIP4Clip} \cite{luo2022clip4clip}., adapts CLIP features for video-text retrieval by calculating similarity between visual and textual embeddings.
\textit{MetaCLIP 2} \cite{chuang2025meta}, a state-of-the-art multi-modal retriever trained on web-scale multilingual data, used here as a high-performance baseline for textual-visual retrieval.\\
\textbf{Implantation Details. } All experiments are conducted on an NVIDIA RTX A6000 GPU. Videos are initially sampled at 0.5 fps.
For video segmentation, "turning points" are identified using either CLIP \cite{radford2021learning} similarity with a 2nd-percentile threshold, or the KTS \cite{potapov2014category}. To ensure semantic coherence, we enforce a minimum clip duration of 5 minutes and a sub-segment minimum of 1.67 minutes.
For each clip, we use VideoLLaMA-7B \cite{zhang2023videollamainstructiontunedaudiovisuallanguage} to generate multi-grained textual descriptions, while frames are sub-sampled at 0.05 fps for fine-grained grounding.
For VideoQA task, we use Contriever \cite{izacard2021contriever} to first retrieve top-10 relevant timeline descriptions. These, along with their corresponding multi-modal clip representations, are then fed into GPT-4o \cite{openai2023gpt4} for answering.
For summarization task, we feed the complete set of timeline-level descriptions directly into GPT-4o to generate a comprehensive video summary.
For video retrieval task, we use Contriever to index all timeline descriptions, returning the top-k segments most relevant to the query.
\begin{figure*}
    \centering
    \includegraphics[width=1.0\linewidth]{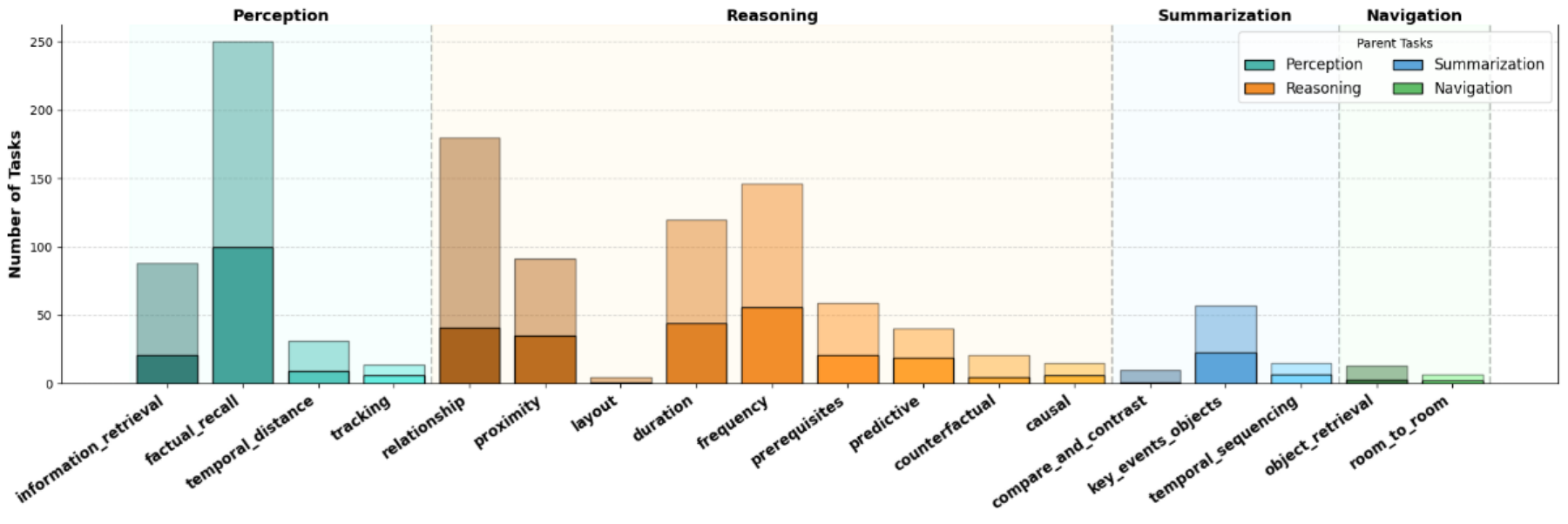}
    \caption{Distribution of Task Categories in Hourvideo Dataset and the Number of Correct Answers by MMViR}
    \label{fig:05}
\end{figure*}
\subsection{Video Question Answering}
Table \ref{tab:qa_comparison} presents a comprehensive comparison of performance between MMViR and baseline models across three long-form VideoQA benchmarks, evaluated on two primary dimensions: \textit{predictive accuracy} and \textit{computational efficiency} (measured by processing time).
As shown, \textbf{MMViR consistently achieves superior QA accuracy across all datasets}. 
\begin{table}[t]
\centering
\footnotesize 
\setlength{\tabcolsep}{3.5pt} 
\begin{tabularx}{\columnwidth}{@{}l ccc ccc@{}}
\toprule
\multirow{2}{*}{\textbf{Method}} & \multicolumn{3}{c}{\textbf{Hourvideo}} & \multicolumn{3}{c}{\textbf{MovieChat-1K}} \\ \cmidrule(lr){2-4} \cmidrule(lr){5-7}
 & \textbf{R-2} & \textbf{R-L} & \textbf{MET.} & \textbf{R-2} & \textbf{R-L} & \textbf{MET.} \\ \midrule
Hourvideo & 10.14 & \textbf{27.33} & 33.87 & 3.31 & 16.54 & 17.16 \\
VideoRecap   & 9.15 & 24.99 & 33.54 & 1.12 & 11.93 & 13.63 \\
MMViR (Ours) & \textbf{10.67} & 27.15 & \textbf{34.98} & \textbf{4.27} & \textbf{18.74} & \textbf{18.52} \\ \bottomrule
\end{tabularx}
\caption{Summarization performance comparison on HourVideo and MovieChat-1K datasets.}
\label{tab:3}
\vspace{-2mm}
\end{table}
\begin{figure}
    \centering
    \includegraphics[width=1.0\linewidth]{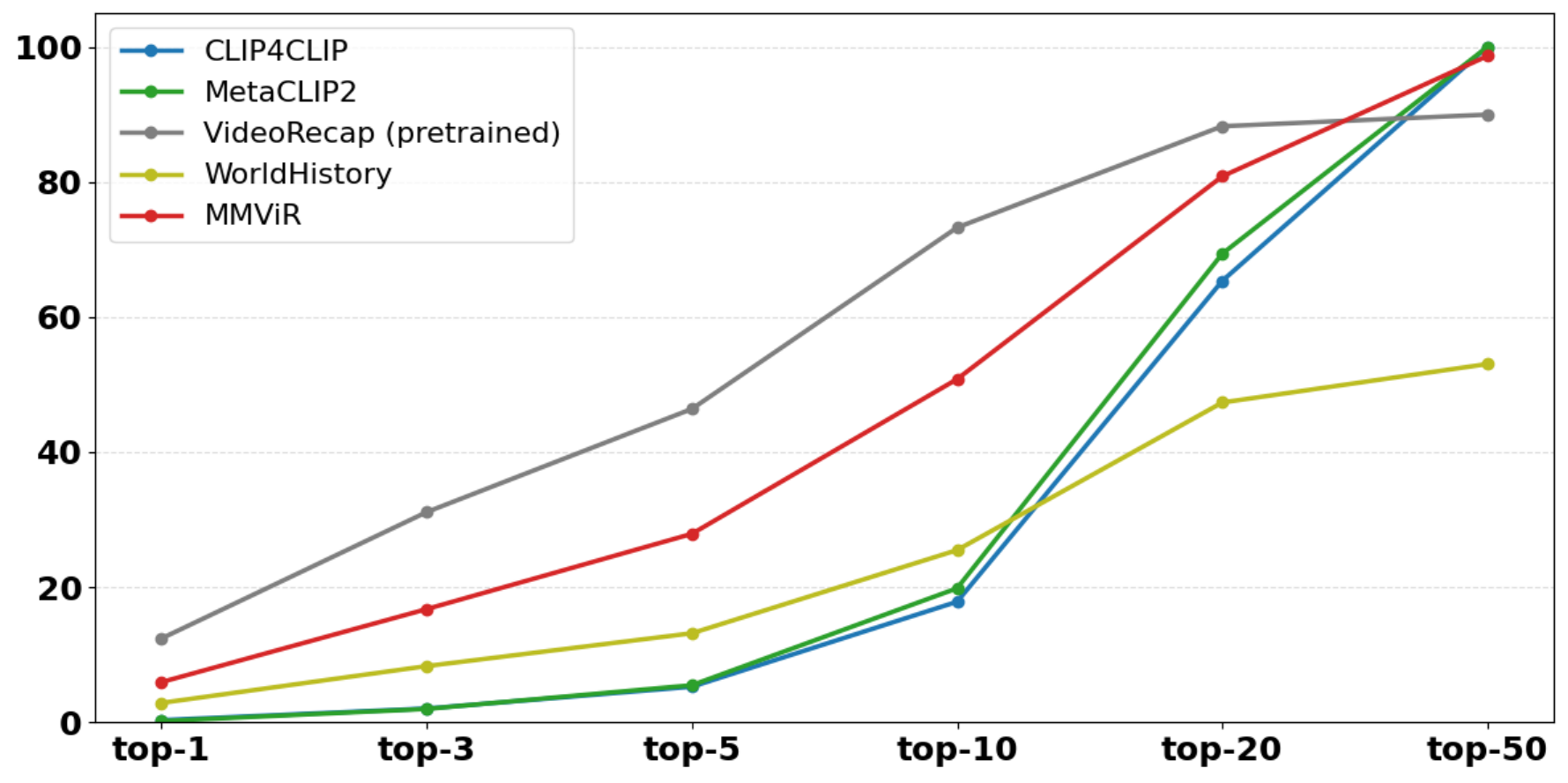}
    \caption{Performance of MMViR and baseline methods on the video retrieval task using \textbf{narration data}.}
    \label{fig:retrieval-1}
    \vspace{-5mm}
\end{figure}
Notably, on the extremely-scale HourVideo benchmark, MMViR achieves the best performance while significantly reducing the processing time compared to competing approaches. 
For datasets with shorter durations and lower scene complexity, such as EgoSchema and Video-MME, we observe that directly retrieving high-precision visual descriptions often yields better results than full multi-modal representations. 
This suggests that for simpler narratives, visual-based scene descriptors provide a more distilled, less noisy context for reasoning.

\begin{figure}
    \centering
    \includegraphics[width=1.0\linewidth]{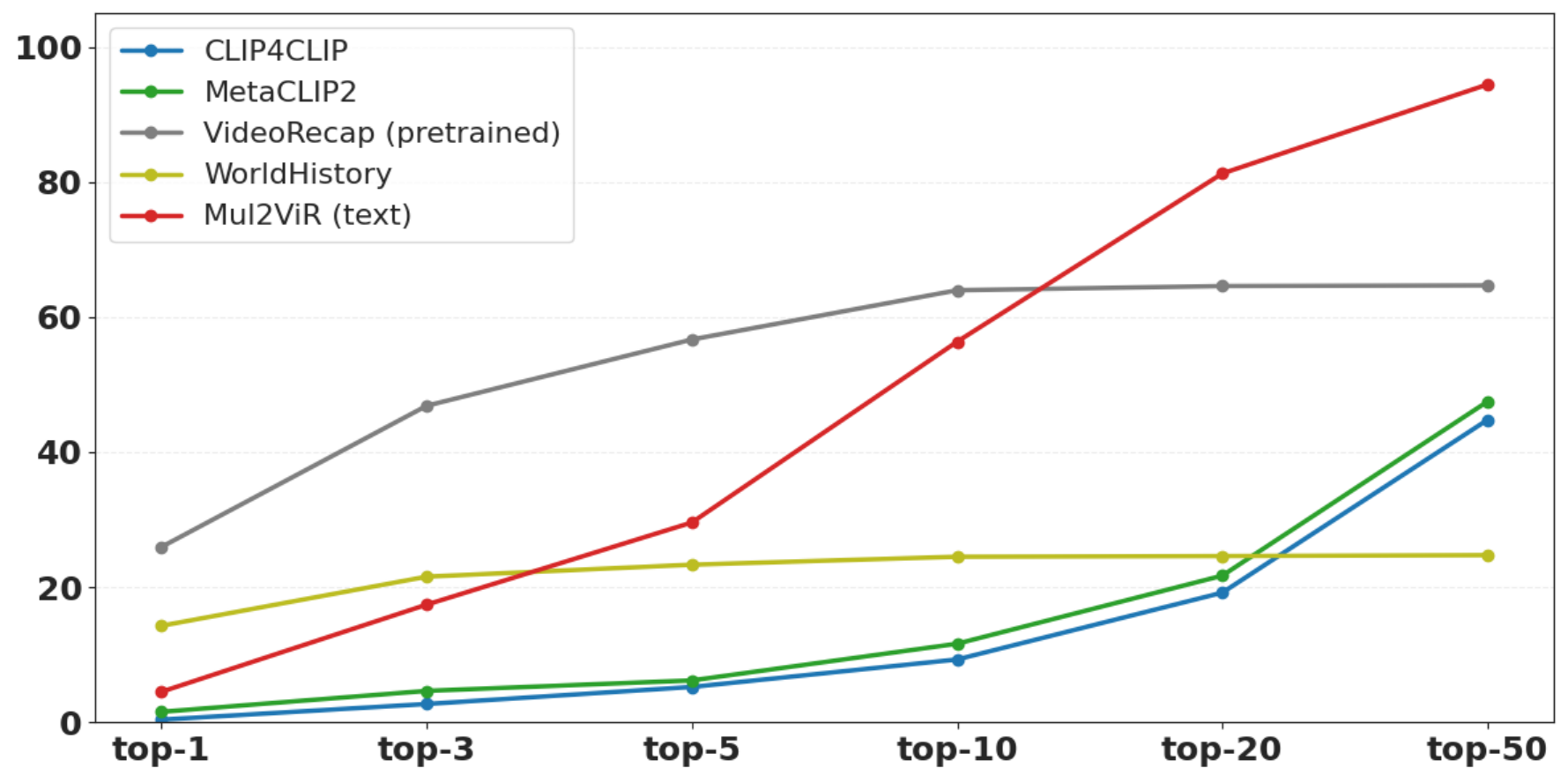}
    \caption{Performance of MMViR and baseline methods on the video retrieval task using \textbf{summary data}.}
    \label{fig:retrieval-2}
    \vspace{-5mm}
\end{figure}

Unlike baseline methods that cause substantial computational overhead due to intensive memory management or exhaustive multi-level captioning, MMViR effectively balances performance and cost. By leveraging an adaptive retrieval mechanism to extract only query-essential information, MMViR demonstrates superior scalability for long-video understanding, effectively bypassing typical computational bottlenecks.

\subsection{Video Summarization} 
Table \ref{tab:3} evaluates the summarization capabilities of MMViR and comparative baselines.
The results indicate that MMViR consistently outperforms competing methods, suggesting that our multi-granularity structure effectively preserves both global narrative coherence and fine-grained semantic details.
While baselines that often struggle with information redundancy or context loss in hour-long videos, MMViR’s timeline-level summaries provide a structured blueprint of the corresponding video.
This allows the LLM to receive comprehensive, logically consistent video abstracts as the input background context, without being overwhelmed by irrelevant temporal noise.
\subsection{Video Retrieval}
Figures \ref{fig:retrieval-1} and \ref{fig:retrieval-2} compare the retrieval performance of MMViR against both text-based and multi-modal baselines.
Notably, MMViR outperforms all non-pretrained baselines, with Top-50 accuracy even surpassing VideoReCap, a model explicitly pretrained on the Ego4D domain.
\begin{figure*}
    \centering
    \includegraphics[width=1.0\linewidth]{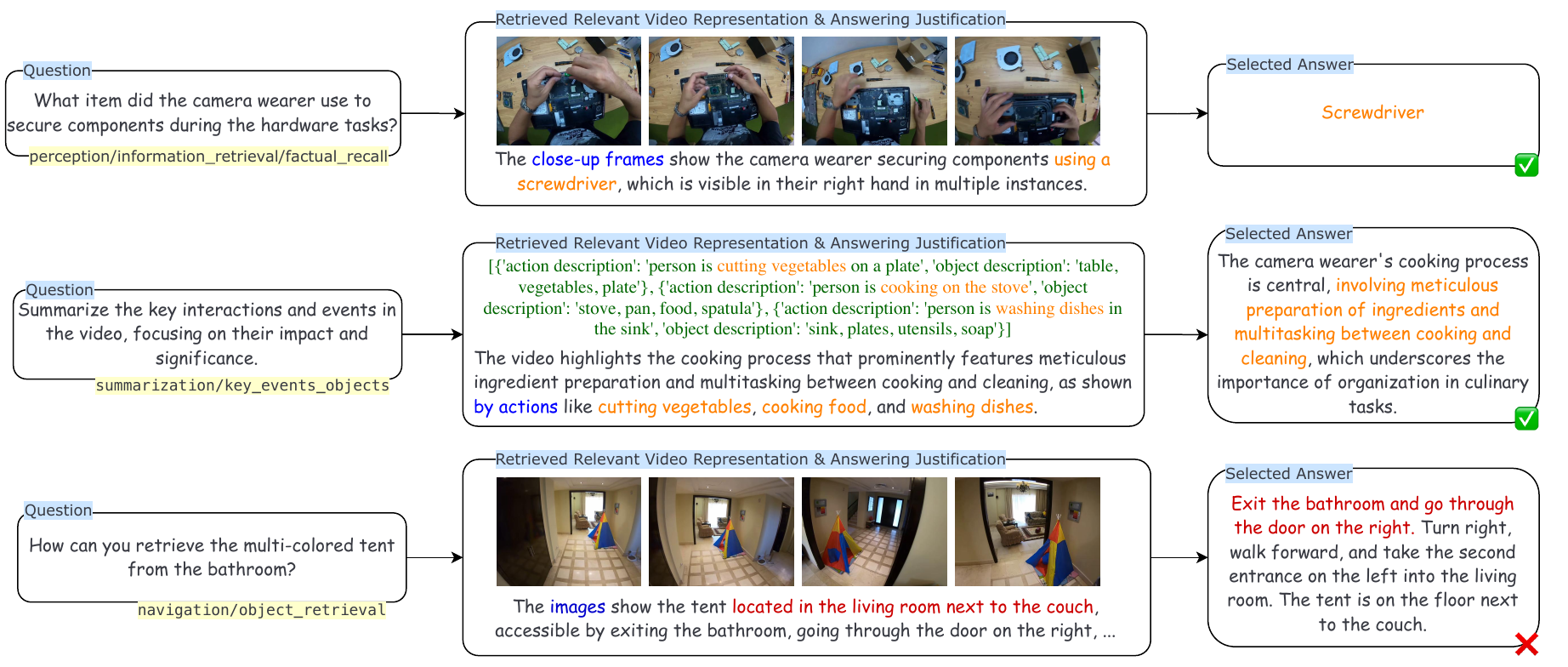}
    \caption{Case study of VideoQA reasoning on HourVideo. Text highlighted in orange/red indicates the logic supporting correct/incorrect answering. The blue-highlighted text indicate the reference modal retrieved from MMViR’s representations.}
    \label{fig:case}
\end{figure*}
The performance gap is obvious in the lower top-$k$ ranges (e.g., $K<10$), where end-to-end multi-modal models like MetaCLIP2 and CLIP4Clip struggle to achieve high precision. This pattern suggests that directly matching queries to raw visual frames in unconstrained long videos remains challenging due to visual redundancy and semantic gaps. In contrast, MMViR’s success underscores the superiority of structured, multi-level textual representations, which serve as a semantic bridge that simplifies complex cross-modal mapping into a more manageable text-to-text alignment task.

\section{Analysis and Discussion}
\subsection{Performance Analysis Across Different Task Categories} 
To further investigate the specialized capabilities of MMViR, we analyze its performance across diverse task sub-types within the HourVideo dataset.
As shown in Figure \ref{fig:05}, MMViR demonstrates robust proficiency across \textit{Perception}, \textit{Reasoning}, and \textit{Summarization} parent tasks, consistently achieving accuracies above 0.4. 
In particular, MMViR excels in \textit{factual recall} in \textit{Perception}, \textit{causal} in \textit{Reasoning}, and \textit{key event object} in \textit{Summarization}.
This performance validates that MMViR's multi-grained representation effectively preserves global narrative coherence while maintaining the semantic depth necessary for complex multi-hop reasoning.

However, performance remains constrained in \textit{Navigation} (avg. acc. 0.25). This discrepancy suggests that while a retrieval-centric framework is highly effective for semantic-based understanding, \textbf{it encounters a "bottleneck" in precise spatio-temporal grounding and explicit geometric relational modeling}. These findings highlight a direction for future studies to integrate a more robust spatial-aware module into long-video reasoning.
\subsection{Case Analysis} 
Figure \ref{fig:case} presents representative examples of MMViR's reasoning process on the HourVideo dataset.
For \textit{Perception} task, MMViR demonstrates a precise ability to anchor its judgments on specific retrieved frames that reveal critical evidence for answering. 
Meanwhile, for higher-level \textit{Summarization} tasks that require broad contextual synthesis, the relevant timeline-level descriptions function as a semantic backbone.
By bridging temporal gaps that discrete frames cannot present, these textual summaries provide a continuous narrative flow, allowing the model to maintain global coherence across extreme durations.
Despite these strengths, the MLLM sometimes misinterprets the background context provided by MMViR due to its intrinsic bottlenecks in cross-modal integration, leading to inaccurate responses. This highlights a future direction to optimize a simpler yet effective representation for MLLM to better understand.

\section{Conclusion}
In this paper, we introduce MMViR, a multi-modal, multi-grained representation designed for long-video understanding. 
By integrating fine-grained visual information with structured textual descriptions, MMViR effectively addresses the challenges of information redundancy and computational constraints.
Extensive experiments validate its robustness across diverse tasks, offering a promising paradigm for future research into efficient, long-range video comprehension.

\section*{Limitations}
While MMViR demonstrates robust performance across long-video benchmarks, several limitations remain for future investigation.
First, our fine-grained visual representations rely on uniform sampling of frames. 
Although this ensures representation completeness, it inevitably introduces temporal redundancy, especially in retrieval tasks where many sampled frames may contain static or redundant information. 

Second, our retrieval mechanism adopts a top-down approach, primarily leveraging high-level timeline summaries to localize relevant segments. 
Although computationally efficient, this strategy may occasionally suffer from a granularity mismatch, potentially overlooking subtle, fine-grained visual information that is not explicitly captured in the textual summaries. 
Future research could explore content-aware adaptive sampling to prioritize informative frames. In addition, developing more advanced cross-modal indexing structures that allow for bi-directional interaction between global narratives and local visual details, to further refine the precision and scalability of long-video comprehension.

\bibliography{custom}

\newpage
\appendix

\section{Clip-based Video Segmentation Analysis}
\begin{figure*}
    \centering
    \includegraphics[width=1.0\linewidth]{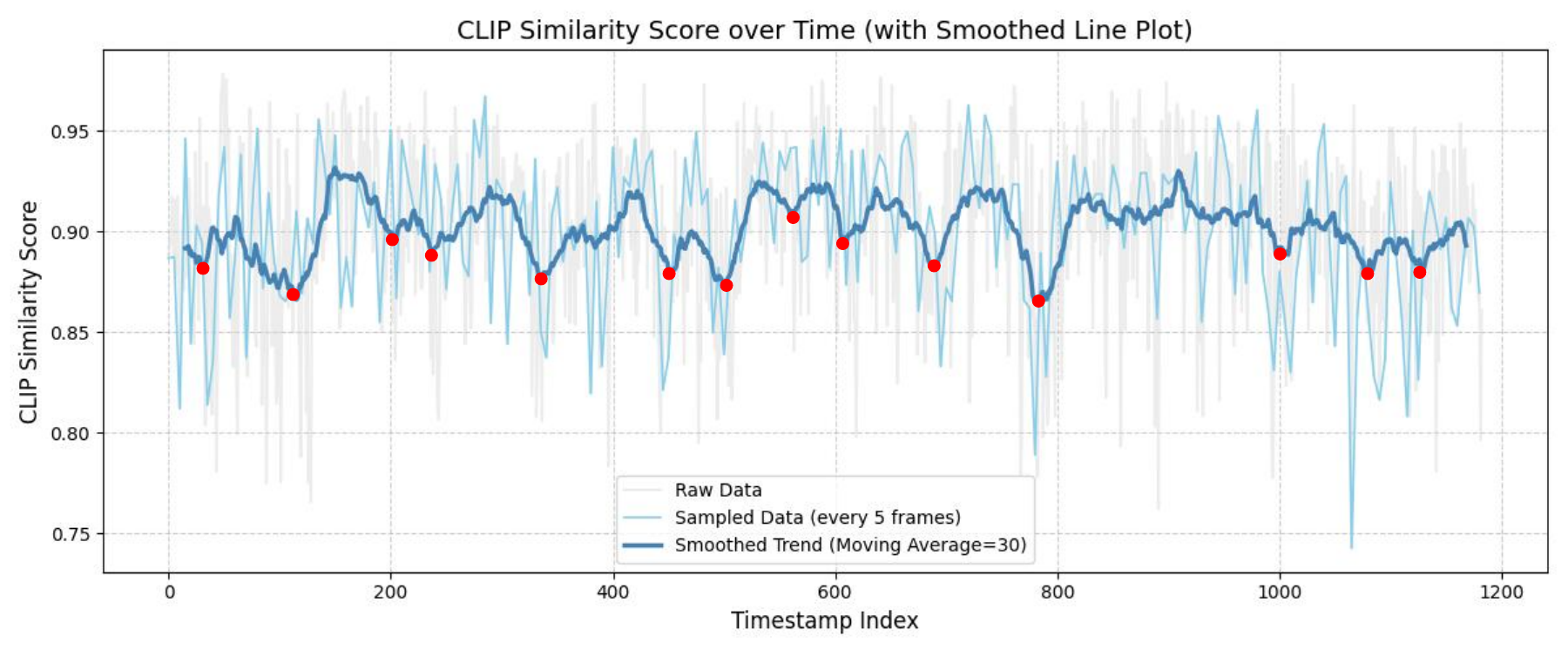}
    \caption{The distribution of CLIP similarity scores over the timestamps and its smoothed curve. The red markers indicate timestamps where noticeable turning points occur.}
    \label{fig:clip-distribution}
\end{figure*}

Beyond the analysis of the overall distribution of frame-level CLIP similarity scores in Section \ref{sec:3}, we further examine the distribution of CLIP similarity scores over the timestamps within an entire video. As illustrated in Figure \ref{fig:clip-distribution}, the local minima (troughs) of the similarity curve are not clustered but are dispersed relatively uniformly across the video timeline. This observation provides additional empirical justification for employing a lower quantile of the CLIP similarity distribution (e.g., the 2nd-percentile) as the threshold for video segmentation.

\label{sec:appendixA}

\section{Impact of Retrieval Scale}
\label{sec:appendixB}
\begin{figure}
    \centering
    \includegraphics[width=1.0\linewidth]{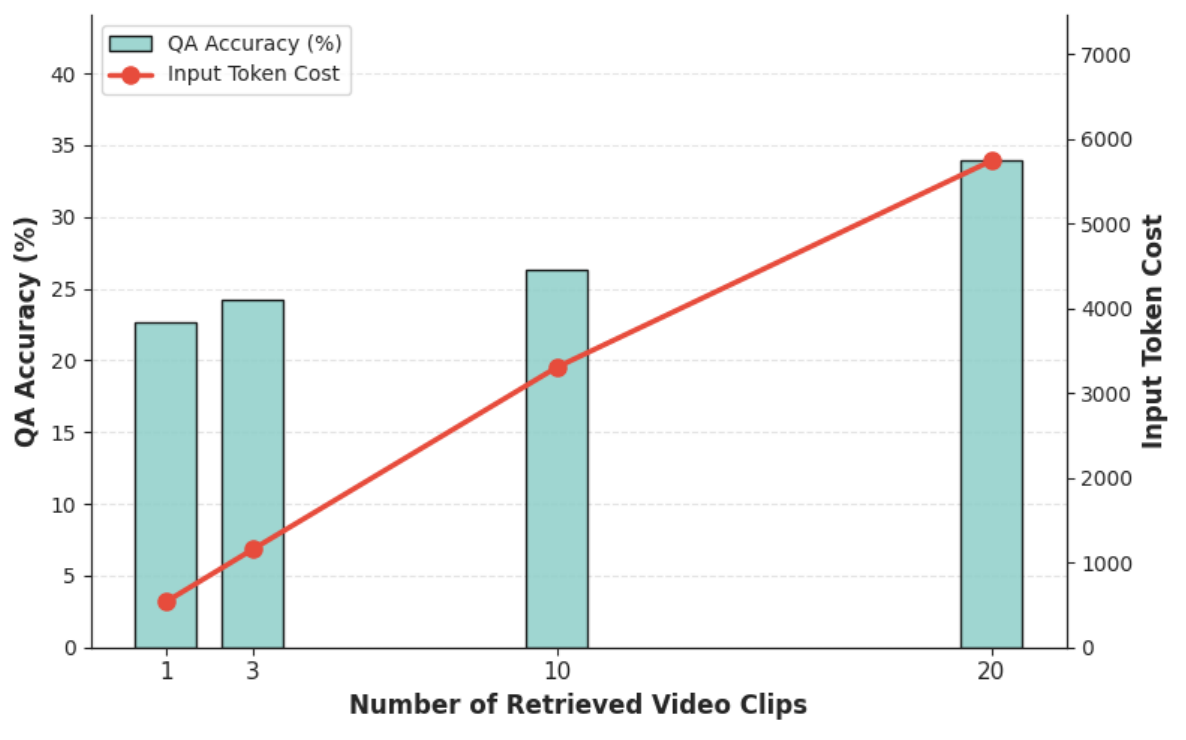}
    \caption{Impact of the Number of Retrieved Clip Representations on QA Accuracy and Inference Cost.}
    \label{fig:06}
\end{figure}

Figure \ref{fig:06} investigates the sensitivity of VideoQA performance to the number of retrieved video clips and their corresponding representations within MMViR.
As shown, retrieving a larger set of relevant clips provides richer contextual cues, thereby consistently enhancing QA accuracy. 
However, this performance gain comes at the cost of increased inference latency and computational overhead, as a higher volume of retrieved descriptions substantially increases the LLM's input token size.
This observation underscores a fundamental trade-off between accuracy and efficiency in long-video understanding. Consequently, the use of an adaptive retrieval strategy is essential to maintain high performance while ensuring computational scalability, allowing the model to selectively process query-critical information without incurring prohibitive token costs.

\section{Instruction Prompt}
\label{sec:appendixC}
In this section, we provide the detailed prompts used to generate the textual representations for MMViR. Following the the descriptive dimensions defined in the HourVideo dataset \cite{chandrasegaran2024hourvideo}, we employ the following prompts to construct the textual video representations: 
\begin{tcolorbox}[
    colback=gray!5, 
    colframe=gray!75, 
    title=High-level Timeline Description, 
    fonttitle=\bfseries,
    arc=2pt, 
    boxrule=0.5pt,
    left=5pt, right=5pt, top=5pt, bottom=5pt
]
\small
\begin{minipage}{0.9\textwidth}
\ttfamily \small
\begin{verbatim}
Your task is to analyze the given video 
frame sequence extracted from a long video 
for a detailed video understanding exercise, 
focusing on the **motion**, to identify and 
describe all of the **actions** appear in the 
images, and **where** whey take place by **who**, 
as well as the corresponding objectives.
YOU ARE ALLOWED TO USE A MAXIMUM OF 50 words for 
this description. Please only return the
description.
\end{verbatim}
\end{minipage}
\end{tcolorbox}

\begin{tcolorbox}[
    colback=gray!5, 
    colframe=gray!75, 
    title=Coarsed-grained (Action Sequence Description), 
    fonttitle=\bfseries,
    arc=2pt, 
    boxrule=0.5pt,
    left=5pt, right=5pt, top=5pt, bottom=5pt
]
\small
\begin{minipage}{0.9\textwidth}
\ttfamily \small
\begin{verbatim}
Your task is to analyze video frames extracted
from a video for a detailed understanding.
I will provide a frame sequence with each frame 
spaced every 2 seconds. Examine these frames 
closely and generate a comprehensive caption 
by strictly following:
List the sequence of all **actions** and the
corresponding **objects** in the order they 
occur in the given frames. 
YOU ARE ALLOWED TO USE A MAXIMUM OF 200 words for 
this description. 
PLEASE Strictly return your results by a list of 
dict in JSON format, following the example below: 
[{'action description': cooking the sausages 
and eggs}, {'action description': cleaning up 
dishes}]
Please do not return a empty list as the result. 
If there is no action appears in the given frames, 
please return the string: 'no action detected'.
\end{verbatim}
\end{minipage}
\end{tcolorbox}

\begin{tcolorbox}[
    colback=gray!5, 
    colframe=gray!75, 
    title=Coarse-grained (Scene Sequence Description), 
    fonttitle=\bfseries,
    arc=2pt, 
    boxrule=0.5pt,
    left=5pt, right=5pt, top=5pt, bottom=5pt
]
\small
\begin{minipage}{0.9\textwidth}
\ttfamily \small
\begin{verbatim}
Your task is to analyze video frames extracted 
from a video for a detailed understanding. 
I will provide a frame sequence with each frame 
spaced every 2 seconds. Examine these frames 
closely and generate a comprehensive caption by 
strictly following: 
List all of the actions and their corresponding 
**settings** that appear in the given frames. 
YOU ARE ALLOWED TO USE A MAXIMUM OF 200 words for 
this description. 
PLEASE Strictly return your results by a list of 
dict in JSON format, following the example below: 
[{'description': The man opens a cabinet in the 
kitchen, 'setting': Kitchen, 'action': Opening 
a cabinet}, {'description': A person is watering 
plants in a garden, 'setting': Garden, 'action': 
watering}]
Please do not return a empty list as the result. 
If there is no action appears in the given frames, 
please return the string: 'no action detected'.
\end{verbatim}
\end{minipage}
\end{tcolorbox}
\begin{tcolorbox}[
    colback=gray!5, 
    colframe=gray!75, 
    title=Coarse-grained (Object Description), 
    fonttitle=\bfseries,
    arc=2pt, 
    boxrule=0.5pt,
    left=5pt, right=5pt, top=5pt, bottom=5pt
]
\small
\begin{minipage}{0.9\textwidth}
\ttfamily \small
\begin{verbatim}
Your task is to analyze video frames extracted 
from a video for a detailed understanding. 
I will provide a frame sequence with each frame 
spaced every 2 seconds. Examine these frames 
closely and generate a comprehensive caption 
by strictly following: 
List the key objects and the characters that 
appear in the given frames, along with their 
attributes if applicable (e.g., color, shape, 
texture), each attribute is separated with a 
comma. 
YOU ARE ALLOWED TO USE A MAXIMUM OF 200 words for 
this description. 
PLEASE Strictly return your results by a list of 
dict in JSON format, following the example below: 
[{'object_name': man, 'number': 1}, 
{'object_name': dog, 'number': 2, 'attributes': 
yellow}]
Please do not return a empty list as the result. 
If there is no object appears in the given frames, 
please return the string: 'no object detected'.
\end{verbatim}
\end{minipage}
\end{tcolorbox}

\begin{tcolorbox}[
    colback=gray!5, 
    colframe=gray!75, 
    title=Fine-grained (Spatial Description), 
    fonttitle=\bfseries,
    arc=2pt, 
    boxrule=0.5pt,
    left=5pt, right=5pt, top=5pt, bottom=5pt
]
\small
\begin{minipage}{0.9\textwidth}
\ttfamily \small
\begin{verbatim}
Your task is to analyze the given image a 
detailed understanding. 
Please examine it closely and generate a 
comprehensive caption by strictly following:
Observe the key objects in the image, and state 
the spatial realtionships between them. List all 
of key objects that appear in the image, along 
with their relationships with others if 
applicable, each relationship is separated with 
a comma. 
YOU ARE ALLOWED TO USE A MAXIMUM OF 200 words for 
this description. 
PLEASE Strictly return your results by a list of 
dict in JSON format, following the example below:
[{'object_name': table, 'number': 1}, 
{'object_name': kettle, 'number': 1, 
'attributes': gray, 'spatial_relationship': 
[on the table, right of the kitchen]}] 
Please do not return a empty list as the result. 
If there is no object appears in the given frames, 
please return the string: 'no object detected'.
\end{verbatim}
\end{minipage}
\end{tcolorbox}

\end{document}